\documentclass[preprint,12pt]{elsarticle}

\usepackage{url}
\usepackage{amssymb}
\usepackage{amsmath}
\usepackage{hyperref}
\usepackage{booktabs}
\usepackage[utf8]{inputenc}
\usepackage{multirow}
\usepackage[table]{xcolor}
\usepackage{bbm}
\usepackage{bm}

\journal{Nuclear Physics B}

\begin{document}

\begin{frontmatter}

\title{UniVector: Unified Vector Extraction via Instance-Geometry Interaction}

\author[1]{Yinglong~Yan}

\ead{yanyl@hnu.edu.cn}

\affiliation[1]{organization={School of Artificial Intelligence and Robotics, Hunan University},
            city={Changsha},
            postcode={410082}, 
            country={China}}

\author[2]{Jun~Yue}

\ead{junyue@csu.edu.cn}

\affiliation[2]{organization={School of Automation, Central South University},
           city={Changsha},
           postcode={410083}, 
           country={China}}
\author[3]{Shaobo~Xia}

\ead{shaobo.xia@csust.edu.cn}

\affiliation[3]{organization={Department of Geomatics Engineering, Changsha University of Science and Technology},
            city={Changsha},
            postcode={410114}, 
            country={China}}
\author[1]{Hanmeng~Sun}

\ead{sunhanmeng7@gmail.com}

\author[1]{Tianxu~Ying}

\ead{yingtianxu3@gmail.com}

\author[1]{Chengcheng~Wu}

\ead{FringsMatalavage@gmail.com}

\author[1]{Sifan~Lan}

\ead{lansifan003@gmail.com}

\author[1]{Min~He}

\ead{hemin@hnu.edu.cn}

\author[4,5]{Pedram~Ghamisi}

\ead{p.ghamisi@gmail.com}

\affiliation[4]{organization={Helmholtz-Zentrum Dresden-Rossendorf (HZDR), Helmholtz Institute Freiberg for Resource Technology, Machine Learning Group},
            city={Freiberg},
            postcode={09599}, 
            country={Germany}}
\affiliation[5]{organization={Lancaster University},
            city={Lancaster},
            postcode={LA1 4YR}, 
            country={U.K.}}
\author[1]{Leyuan~Fang\corref{cor1}}


\ead{fangleyuan@gmail.com}

\cortext[cor1]{Corresponding author.}


\begin{abstract}
Vector extraction (VE) retrieves structured vector geometry from raster images, offering high-fidelity representation and broad applicability. Existing methods, however, are usually tailored to a single vector type (e.g., polygons, polylines, line segments), requiring separate models for different structures. This stems from treating instance attributes (category, structure) and geometric attributes (point coordinates, connections) independently, limiting the ability to capture complex structures. Inspired by the human brain’s simultaneous use of semantic and spatial interactions in visual perception, we propose UniVector, a unified VE framework that leverages instance–geometry interaction to extract multiple vector types within a single model. UniVector encodes vectors as structured queries containing both instance- and geometry-level information, and iteratively updates them through an interaction module for cross-level context exchange. A dynamic shape constraint further refines global structures and key points. To benchmark multi-structure scenarios, we introduce the Multi-Vector dataset with diverse polygons, polylines, and line segments. Experiments show UniVector sets a new state of the art on both single- and multi-structure VE tasks. Code and dataset will be released at \href{https://github.com/yyyyll0ss/UniVector}{https://github.com/yyyyll0ss/UniVector}.
\end{abstract}




\end{frontmatter}


\section{Introduction}\label{sec:introduction}
\begin{figure}[!t]
	\centering
	\includegraphics[width=1.0\textwidth]{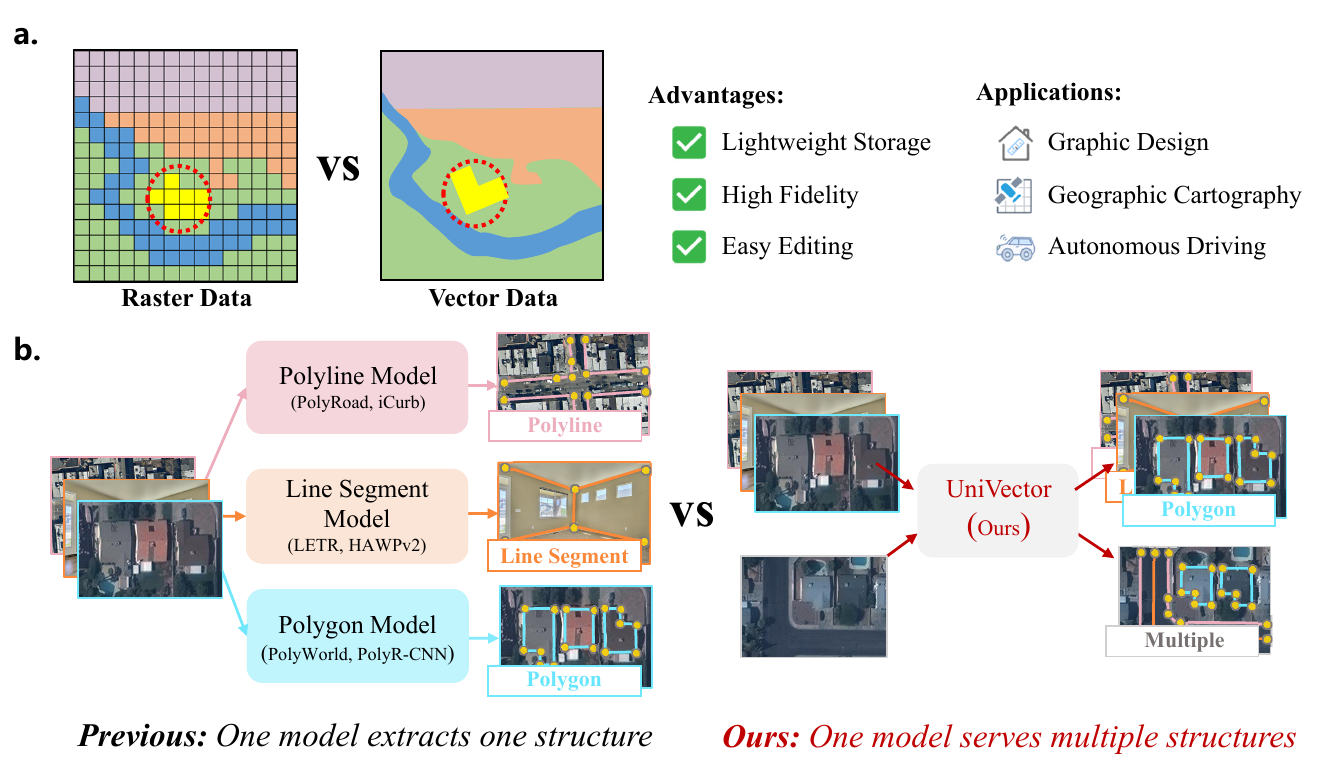}
	\caption{\textbf{a.} Compared with raster images, vector data are lightweight, high-fidelity, and easily editable, and are widely used in graphic design, geographic cartography, and autonomous driving. \textbf{b.} Comparison of specific vs. unified vector extraction: while prior models \cite{xue2023holistically,zorzi2022polyworldg,jiao2024polyrcnn,hu2024polyroad,xu2021icurb,xu2021line} handle only one vector structure, UniVector extracts multiple structures within a single framework.}
	\label{fig_1}
\end{figure}

Vector information serves as a fundamental cognitive unit of visual perception, enabling accurate representation of spatial properties of the physical world \cite{bicanski2020neuronal}, \cite{zhao2022deep}, \cite{10194945}, such as location, shape, and layout. Vector extraction (VE) is a core computer vision task that retrieves structured vector information from raster images, and vector data offers lightweight storage, high fidelity, and easy editability compared with raster data (as shown in Fig. \ref{fig_1}a). With advances in imaging technology, high-definition large-scale images can now be acquired, covering diverse objects and structures, including building contours \cite{zorzi2022polyworldg}, \cite{xu2022accurate}, \cite{jiao2024polyrcnn}, road networks \cite{xue2022quantifying}, \cite{he2020sat2graph}, road boundaries \cite{hu2024polyroad}, \cite{xu2021icurb}, and wireframes \cite{xue2023holistically}, \cite{xu2021line}. Therefore, accurately extracting multiple vector structures in large-scale images is essential for various applications, including graphic design \cite{yue2023connecting}, geographic cartography \cite{xue2022quantifying}, \cite{10374326}, and autonomous driving \cite{liao2024maptrv2}.

Vector extraction (VE) requires modeling both instance-level structure and fine-grained geometry. Existing approaches typically decompose VE into two cascaded sub-tasks and can be grouped into two paradigms: instance-to-geometry and geometry-to-instance. (1) Instance-to-geometry methods \cite{xu2022accurate,jiao2024polyrcnn} first predict instance representations (e.g., bounding boxes or masks) and then generate geometric shapes, leveraging advances in segmentation and detection \cite{he2017mask,carion2020end}. These methods are straightforward but depend heavily on instance quality and may distort complex shapes such as elongated polylines. (2) Geometry-to-instance methods \cite{xue2023holistically,zorzi2022polyworldg} detect geometric points first and infer their connections, yielding more accurate shapes and better scalability \cite{wang2023regularized}. However, the lack of instance-level constraints often causes topology errors in multi-structure scenes. Most existing techniques are tailored to specific vector types, requiring separate models for different structures \cite{wang2023regularized}, as illustrated in Fig.~\ref{fig_1}b. Achieving unified vector extraction (UVE) across diverse structures therefore remains a key challenge.

\begin{figure*}[t]
	\centering
	\includegraphics[width=1\textwidth]{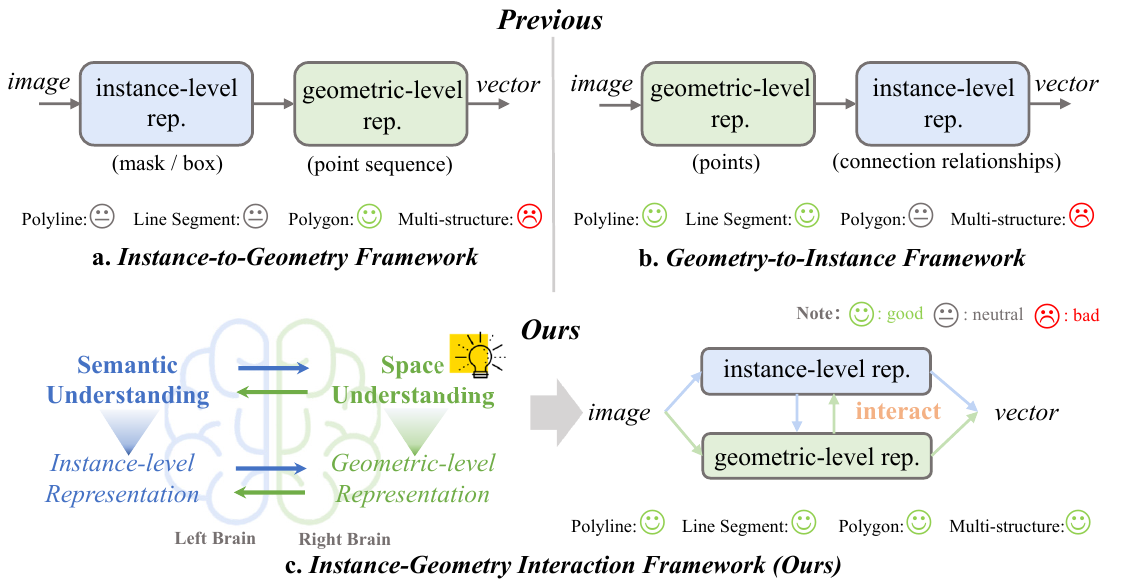}
	\caption{\textbf{Comparison of previous frameworks and our UniVector.} Existing methods \cite{zorzi2022polyworldg}, \cite{xu2022accurate}, \cite{jiao2024polyrcnn} usually split vector extraction into two cascaded tasks, often causing shape inaccuracies or topological errors. Inspired by the human brain’s simultaneous use of semantic and spatial interactions in visual perception, UniVector models instance–geometry interaction to capture both precise shapes and topology across diverse structures.}
	\label{fig_2}
\end{figure*}

Previous methods \cite{jiao2024polyrcnn,wang2023regularized} typically follow a cascaded pipeline, modeling vector instances and geometric attributes separately and ignoring the information gap between them. As shown in Fig.~\ref{fig_2}(a, b), vectors naturally contain instance-level attributes (semantic category, structural connectivity) and geometric-level attributes (point coordinates and connections) \cite{zorzi2022polyworldg}. Using only instance cues fails to capture precise shapes, while relying solely on geometry cannot guarantee correct topology. A joint representation of both, however, accurately describes diverse structures. Thus, cascaded approaches (Fig.~\ref{fig_2}(a, b)) limit the ability to learn complex vector forms. The human brain often relies on the interplay of semantic and spatial understanding for visual perception, with the two processes occurring simultaneously and mutually reinforcing each other. Inspired by this, we model explicit instance–geometry interaction (Fig.~\ref{fig_2}(c)) to bridge this gap, allowing global structural priors from instance attributes and fine semantic–structural cues from geometry to complement each other

In this paper, we propose UniVector, a unified framework that encodes different vectors into a shared representation and dynamically refines their positions and shapes through instance-geometry interaction. First, we introduce a unified vector encoder, which converts common instance-geometric attributes (\emph{e.g.}, category, structure, position, shape) into structured queries that serve as interactive learning carriers. To facilitate parallel interactions between instance and geometric features, we design an instance-geometry interaction decoder that iteratively refines these queries, reducing single-level information bias and achieving coherent feature integration. Additionally, we develop a Dynamic Shape Constraint (DSC) to adaptively balance global structural consistency and local shape accuracy, significantly enhancing performance in complex scenarios.

Existing VE datasets \cite{xu2021topoboundary,mohanty2018crowdai} cover only single vector types. We therefore build Multi-Vector, the first dataset for multi-structure VE, comprising polygons, polylines, and line segments across three semantic categories—buildings, road boundaries, and centerlines—with 20,000 training and 3,734 test images. Experiments show UniVector achieves state-of-the-art performance on both single- and multi-structure VE tasks. Our main contributions are:
\begin{itemize}
\item{\textbf{Unified Representation \& Framework:} We propose a structured query representation for various vector structures and introduce UniVector, an instance-geometry interaction learning framework for unified vector extraction (UVE).}

\item{\textbf{Instance-Geometry Interaction Modeling:} We design a unified vector encoder and an instance-geometry interaction decoder to adaptively initialize and refine structured queries.}

\item{\textbf{Dynamic Shape Constraint (DSC):} To address shape discrepancies across different vectors, we introduce DSC, which dynamically optimizes both global structure consistency and local shape accuracy.}

\item{\textbf{Multi-Vector Dataset:} To validate our approach, we construct the first multi-structure VE dataset (Multi-Vector) containing polygons, polylines, and line segments. Our method consistently outperforms existing approaches in both specific-structure and multi-structure VE tasks.}
\end{itemize}

\section{Related Work}
\subsection{Different Structures in Vector Extraction}

Raster images contain rich vector information, typically organized into three basic geometric structures: polygons \cite{zorzi2022polyworldg,zhang2022e2ec}, polylines \cite{he2020sat2graph,tan2020vecroad}, and line segments \cite{xue2023holistically,zhao2022deep}, each conveying distinct geometric and semantic characteristics.

\textbf{Polygons.} Defined by a closed point sequence, polygons outline object contours and are widely used for building extraction \cite{zorzi2022polyworldg,xu2022accurate,jiao2024polyrcnn}, contour-based instance segmentation \cite{zhang2022e2ec}, and high-definition mapping \cite{liao2024maptrv2}. Their adjustable vertex count supports targets of varying complexity.

\textbf{Polylines.} With open, directed topology, polylines effectively represent linear structures such as road boundaries \cite{xu2021icurb} and lanes \cite{li2023lanesegnet}; complex road networks are often modeled as combinations of polylines \cite{he2020sat2graph}.

\textbf{Line segments.} As the most basic vector units, line segments are essential for wireframe parsing \cite{xue2023holistically,huang2018learning} and semantic line detection \cite{zhao2022deep}, and capture regular edges in man-made environments.

Despite significant progress, most methods \cite{xue2023holistically,zorzi2022polyworldg,he2020sat2graph} focus on a single vector type, overlooking geometric relationships—e.g., polygons and polylines both consist of multiple line segments—thus requiring multiple models and raising deployment cost. To overcome these limits, we present the Multi-Vector dataset and UniVector, a unified approach for efficient, cross-structure vector extraction.

\subsection{Mainstream Methods in Vector Extraction}

Vector extraction (VE) has long been a challenge in computer vision. Early methods relied on hand-crafted low-level cues—such as gradients \cite{canny1986computational} and textures \cite{huang2018learning}—but their heuristic nature often caused large errors and suboptimal results. Consequently, research shifted to deep learning–based approaches \cite{jiao2024polyrcnn,he2020sat2graph}, which model both instance-level and geometric attributes, typically through cascaded architectures. This section reviews the two prevailing paradigms: instance-to-geometry and geometry-to-instance.

\subsubsection{Instance-to-Geometry Framework}
Instance-to-geometry methods \cite{xu2022accurate,girard2021polygonal,liang2019convolutional} first predict instance representations (e.g., boxes or masks) and then infer vector geometry. Early approaches leveraged semantic segmentation \cite{he2017mask,cheng2022maskedattention}: for example, building masks were simplified into polygons via Douglas–Peucker \cite{douglas1973algorithms}, Frame Field Learning combined frame fields with masks \cite{girard2021polygonal}, and road centerlines were refined from binarized masks \cite{batra2019improved}. Wireframe parsing used junction and line heatmaps merged into vectors \cite{huang2018learning}. While masks offer shape cues, they often fail with overlapping instances.

Later methods replaced masks with instance features from object detection \cite{jiao2024polyrcnn,zhang2024p2pformer}. Castrejon et al. \cite{castrejon2017annotating} applied RNNs to sequentially predict polygon vertices. Xu et al. \cite{xu2024patched} reconstructed roads by merging learned line segments. Inspired by DETR \cite{carion2020end}, LETR \cite{xu2021line} and PolyR-CNN \cite{jiao2024polyrcnn} employed instance queries with iterative cross-attention for point prediction, while P2PFormer \cite{zhang2024p2pformer} refined coordinates through ordered point queries.

Despite these advances, the framework’s serial pipeline makes early errors hard to correct and depends heavily on mask or instance quality, causing distortions in dense scenes and limiting generalization across vector types.

\subsubsection{Geometry-to-Instance Framework} 
The Geometry-to-Instance framework represents vector annotations as a graph, first detecting points and then predicting connections \cite{zorzi2022polyworldg,yang2023topdiga,tan2020vecroad,he2022td}. Early methods, such as RoadTracer \cite{bastani2018roadtracer} and VecRoad \cite{tan2020vecroad}, generated points iteratively, while Rngdet \cite{xu2022rngdet}incorporated contextual features for better accuracy. Later approaches—including TD-Road \cite{he2022td}, PolyWorld \cite{zorzi2022polyworldg}, and GraphMapper \cite{wang2023regularized}—extract all points simultaneously using graph neural networks (GNNs) to predict connections, with enhancements like dense feature sampling, weighted neighbor features and attention-based GNNs. Recent methods \cite{wang2023regularized,zhang2024p2pformer} leverage higher-level primitives (e.g., line segments, angles) to improve efficiency and robustness against occlusion or shadows.

By minimizing heuristic design, Geometry-to-Instance methods achieve strong performance and scalability \cite{yang2023topdiga,wang2023regularized}, and are widely applied in high-definition map construction \cite{liao2024maptrv2}. However, lacking instance-level priors makes distinguishing overlapping instances difficult, leading to topological errors and limiting multi-structure vector extraction.

To overcome these limitations, we propose a unified representation that integrates instance and geometric attributes and models their interaction. Instance-level attributes provide global structural priors for topology and coordinates, while geometric attributes enhance semantic and structural differentiation, enabling complementary advantages.

\section{Method}

\begin{figure*}[t]
	\centering
	\includegraphics[width=1\textwidth]{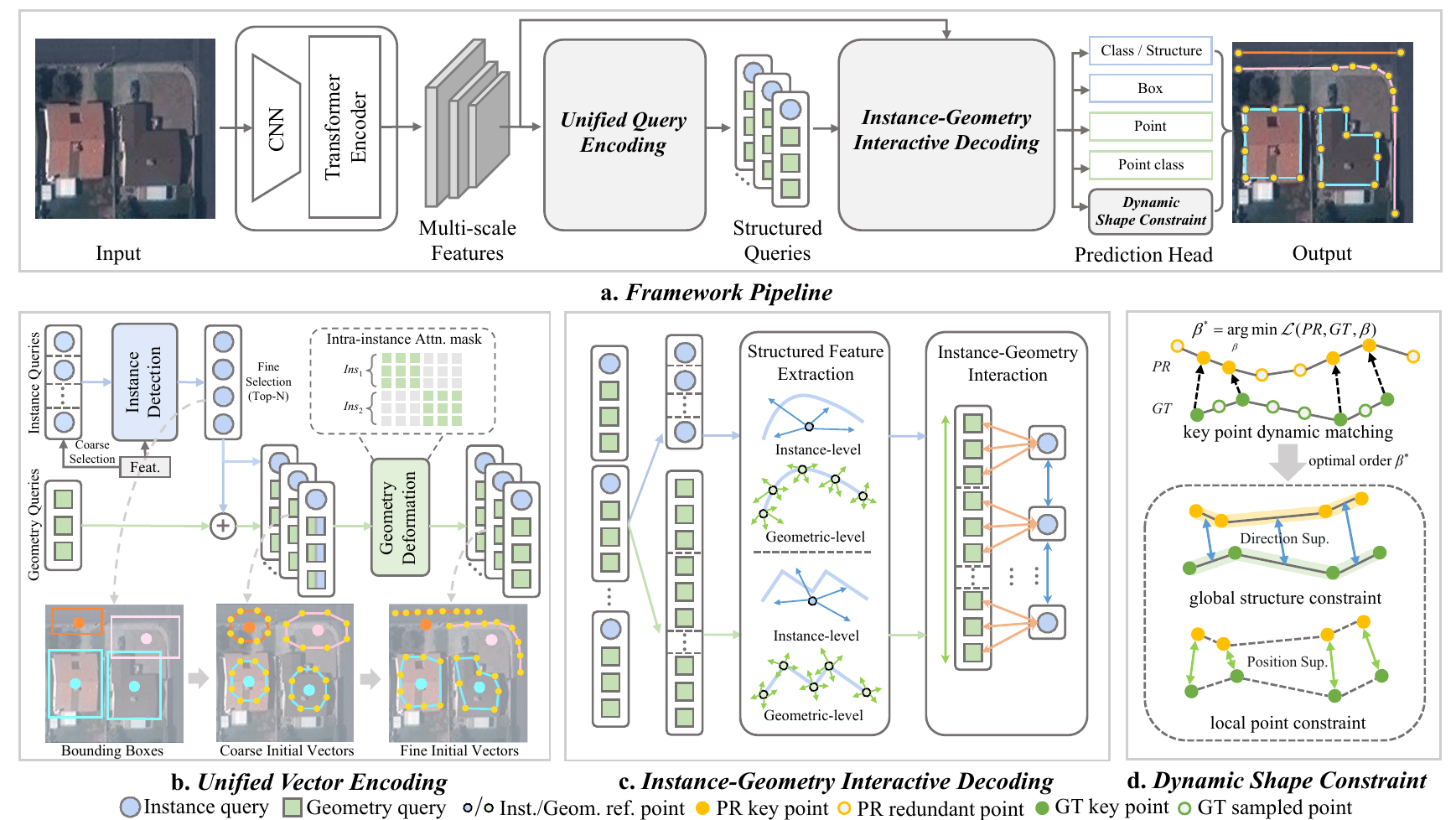}
	\caption{\textbf{Overview of UniVector.} \textbf{a.} The pipeline includes unified vector encoding, instance-geometry interactive decoding, and a dynamic shape constraint. \textbf{b.} Vector instances and geometric attributes are first encoded as unified queries for interactive learning. \textbf{c.} Instance-geometry interaction then iteratively refines these queries for cross-level learning. \textbf{d.} A dynamic shape constraint ensures global structural consistency and local accuracy.}
	\label{fig_3}
\end{figure*}

\subsection{Overall Framework}
The UniVector framework (Fig.~\ref{fig_3}a) comprises three main components: unified vector encoding, instance–geometry interactive decoding, and dynamic shape constraint. A CNN backbone with a Transformer encoder \cite{zhu2020deformable} extracts multi-scale image features $F$, which are encoded into structured queries $Q_s$ by the unified vector encoder. $Q_s$ combines instance queries $Q_{ins}$ and geometric queries $Q_{geo}$, representing instance- and geometry-level information. An instance–geometry interactive decoder iteratively refines $Q_s$, while the dynamic shape constraint ensures global structural consistency and local geometric accuracy. The optimized queries are then processed by prediction heads to generate instance classes, bounding boxes, point coordinates, and point categories—the latter filtering key points for concise shapes, and bounding boxes providing auxiliary supervision for faster convergence.

\subsection{Unified Vector Encoding}
Unified vector extraction requires encoding vectors of different structures into a single representation. Traditional methods represent vectors as masks \cite{xu2022accurate}, graphs \cite{zorzi2022polyworldg}, or point sequences \cite{jiao2024polyrcnn}, but these approaches are often biased toward either instance- or geometry-level attributes, limiting their generality. The key challenge is to encode both attributes simultaneously. Queries provide a flexible medium for representing diverse objects \cite{carion2020end,xu2021line,yue2023connecting}. We introduce structured queries to jointly encode instance- and geometric-level information, treating each vector as a unit represented by its holistic structure and spatial coordinates. This allows vectors to learn their own attributes while interacting with others. In this section, we describe the unified representation using Structured Queries and the Query Encoding process (Fig.~\ref{fig_3}b).

\begin{figure*}[t]
	\centering
	\includegraphics[width=1.0\textwidth]{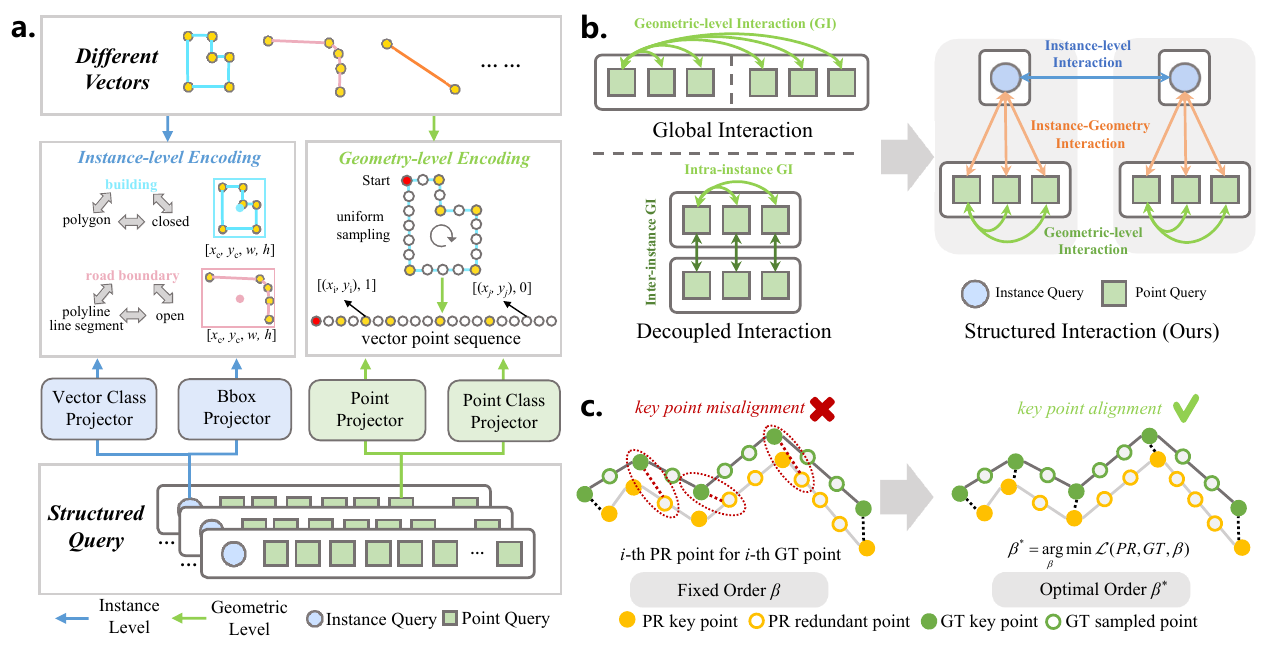}
	\caption{\textbf{a.} The process of encoding vectors into structured queries. \textbf{b.} Comparison of query interaction: the global interaction at geometric-level \cite{yue2023connecting}; the decoupled interaction at geometric-level \cite{liao2024maptrv2}; and our structured instance-geometry interaction. \textbf{c.} Motivation of Dynamic Shape Constraints (DSC). Fixed-order constraints risk keypoint misalignment, while DSC dynamically matches predictions to ground truth for optimal keypoint pairing.}
	\label{fig_4}
\end{figure*}

\textbf{Structured Query.} To capture both instance- and geometry-level information, we encode vectors using structured queries $Q_s \in \mathbb{R}^{N \times (M+1) \times C}$, where $N$, $M$, and $C$ denote the maximum number of vector instances, points per vector, and channel dimensions (Fig.~\ref{fig_4}a). Each vector $Q_s^i \in \mathbb{R}^{(M+1) \times C}$ consists of an instance query $Q_{ins}^i \in \mathbb{R}^{C}$ and a geometric query $Q_{geo}^i \in \mathbb{R}^{M \times C}$, representing instance- and geometry-level attributes, respectively. Instance queries encode semantic categories and structural types, linking class, topology, and geometric openness. Spatial positions are represented via bounding boxes. Geometric queries use uniform sampling to align point sequences, with closed polygons sampled clockwise from the top-left point and open polylines or line segments sampled bidirectionally \cite{liao2024maptrv2}, selecting the sequence with smaller prediction error. This process unifies diverse vector shapes and structures into a consistent structured query set $Q_s$.

\textbf{Query Encoding.} After establishing a unified representation, vector information is encoded into structured queries via instance detection and geometry deformation modules (Fig.~\ref{fig_3}b). For instance-level encoding, instance queries $Q_{ins}$ capture attributes such as categories and bounding boxes. Unlike random initialization \cite{carion2020end} or simple query selection \cite{zhu2020deformable}, we adopt a coarse-to-fine strategy: top-scoring image tokens (\emph{e.g.}, top 900) form coarse queries $Q_{ins}^c$, which are refined in the instance detection module to select the top $N$ queries as $Q_{ins}$. A lightweight two-layer transformer decoder reduces computation while encoding more accurate instance information. For geometric-level encoding, geometry queries $Q_{geo}$ parameterize uniformly sampled point sequences. Coarse geometry queries $Q_{geo}^c$ are generated by summing instance queries $Q_{ins}$ with a learnable embedding $V$, but these lack inter-point correlations. To capture detailed structures, a shape deformation module with intra-instance attention refines $Q_{geo}$ through point-wise interactions:
\begin{equation}
Q_{geo}^{i,j} = f(Q_{geo}^{i,j}, Q_{geo}^{i}) = \sum_{n=1}^M w_{i,j} \, \phi(Q_{geo}^{i,j}),
\end{equation}
where $w_{i,j}$ are learnable weights, $\phi(\cdot)$ a nonlinear transformation, and $f(\cdot)$ denotes self-attention. This aligns point queries with true geometric shapes. Together, these modules encode instance and geometric information into structured queries, initializing the subsequent decoding process.

\subsection{Instance-Geometry Interactive Decoding} After unifying instance and geometric attributes into structured queries, we refine them iteratively to decode the final results. Existing decoders \cite{jiao2024polyrcnn,hu2024polyroad,liao2024maptrv2} mainly target single-level queries—either instance \cite{zhu2020deformable} or point \cite{liao2024maptrv2}—and thus fail to exploit the multi-level context of structured queries. To address this limitation, we propose an instance–geometry interaction decoding strategy that integrates structured feature extraction with cross-level aggregation, enabling orderly fusion of instance and geometric information for progressive refinement.

\textbf{Structured Feature Extraction.}
To extract features at different granularities, we enhance deformable attention \cite{zhu2020deformable} by equipping each vector with instance reference points $R_{ins} \in \mathbb{R}^{N \times 2}$ and geometric reference points $R_{geo} \in \mathbb{R}^{N \times M \times 2}$. The update mechanism for instance reference points is similar to object detection. The geometric reference points in the first layer are derived from the instance reference points through offset learning, and in subsequent layers, they are iteratively updated using the preceding layer’s reference points:

To capture multi-scale features, we extend deformable attention \cite{zhu2020deformable} by assigning each vector instance reference points $R_{ins} \in \mathbb{R}^{N \times 2}$ and geometric reference points $R_{geo} \in \mathbb{R}^{N \times M \times 2}$. Instance references update as in object detection, while geometric references are initialized from $R_{ins}$ via offset learning and iteratively refined using the previous layer's references:
\begin{equation}
	\begin{aligned}
        \begin{cases}
        & R_{geo}^{l} = \mathrm{Sigmoid}(\mathrm{Sigmoid^{-1}}(R_{ins}^l)+\mathrm{MLP}(Q_{geo}^{l})), l=0\\
		& R_{geo}^{l} = \mathrm{Sigmoid}(\mathrm{Sigmoid^{-1}}(R_{geo}^l)+\mathrm{MLP}(Q_{geo}^{l})), l>=1
        \end{cases}
	\end{aligned}
\end{equation}
where $l$ represents the current layer index, $\mathrm{Sigmoid}$ and $\mathrm{Sigmoid^{-1}}$ are the sigmoid and inverse sigmoid activation functions, and MLP refers to a Multi-Layer Perceptron layer. All reference points $R_{s}^l \in \mathbb{R}^{N \times (M+1) \times 2}$ are assigned $E$ sampling points to facilitate effective aggregation of contextual information. The coordinate offsets $\Delta S_{s}^l$ and sampling coordinates $S_{s}^l$ are computed as:
\begin{equation}
	\begin{aligned}
        & \Delta S_{s}^l = \mathrm{Sampling \_ offset}(Q_{s}^{l-1}) \in \mathbb{R}^{N \times (M+1) \times E \times 2}\\
		& S_{s}^l = R_{s}^{l-1}+\Delta S_{s}^l \in \mathbb{R}^{N \times (M+1) \times E\times2},
	\end{aligned}
\end{equation}
where $\mathrm{Sampling\_offset}$ is a linear projector. Subsequently, the structured query $Q_s^l$ is updated through a weighted summation of the sampled features:
\begin{equation}
	\begin{aligned}
        & W_{s}^l = \underset{k \in E} {\mathrm{Softmax}}(W_{j,k}^l) \in \mathbb{R}^{N \times (M+1) \times E}\\
		& Q_{s}^l = \sum_{k=1}^E[W_{s}^l\cdot \textrm{Sampling}(F,S_{j,k}^l)] \in \mathbb{R}^{N \times (M+1) \times C}.
	\end{aligned}
\end{equation}
Here $j$ indexes the $(M+1)$ points of $Q_{s}^l$, $k$ indexes the $E$ sampling points, $W_{s}^l$ is the softmax weights, and Sampling is the bilinear operator. This structured sampling progressively refines with multi-scale features, capturing both global and local vector information more effectively than purely instance- or geometry-based methods \cite{hu2024polyroad,carion2020end,yue2023connecting,liao2024maptrv2} and supporting later interactive learning.

\textbf{Instance-Geometry Interaction.}
After extracting multi-level features, we apply instance–geometry interaction for cross-level complementarity (Fig.~\ref{fig_4}b. Unlike previous global \cite{yue2023connecting} or decoupled \cite{liao2024maptrv2} interactions, we use a structured scheme: cross-level attention links instance queries with all point queries for global guidance, intra-instance attention refines points via neighboring features, and inter-instance interaction enables cross-target learning. For instance queries $Q_{ins}\in \mathbb{R}^{N \times C}$ and geometry queries $Q_{geo}\in \mathbb{R}^{N \times M \times C}$, single-level interaction uses self-attention (SA):
\begin{equation}
	\begin{aligned}
        & Q_{ins}^{'} = \textrm{SA}(Q_{ins})\in \mathbb{R}^{N \times C}\\
		& Q_{geo}^{'} = \textrm{Concat}(\textrm{SA}(Q_{geo}^{i}), i \in [1,...,N])\in \mathbb{R}^{N \times M \times C},
	\end{aligned}
\end{equation}
Cross-level refinement uses cross-attention (CA):
\begin{equation}
	\begin{aligned}
        & Q_{ins}^{''} = \textrm{Concat}(\textrm{CA}(Q_{ins}^{i'}, Q_{geo}^{i'}), i \in [1,...,N])\in \mathbb{R}^{N \times C}\\
		& Q_{geo}^{''} = \textrm{Concat}(\textrm{CA}(Q_{geo}^{i'}, Q_{ins}^{i'}), i \in [1,...,N])\in \mathbb{R}^{N \times M \times C},
	\end{aligned}
\end{equation}
Here, $Q_{ins}^{''}$ and $Q_{geo}^{''}$ are refined queries after cross-level interaction. This allows instance queries to incorporate geometric guidance and geometry queries to benefit from instance semantics. The updated queries are then merged to form the structured query set $Q_{s}^{''}$.

\subsection{Dynamic Shape Constraint (DSC)}
Vector shapes vary greatly, demanding flexible supervision. Fixed vertex pairings (Fig.~\ref{fig_4}c) often misalign when shapes or point counts differ. We address this with point-level dynamic matching \cite{carion2020end}. The proposed Dynamic Shape Constraint (DSC) adaptively pairs predicted and reference points, enforcing both global structure and local accuracy.

\textbf{Key Point Dynamic Matching.}
Previous single-structure VE tasks assumed a fixed number of points per target, enabling only instance-level matching \cite{cheng2022maskedattention, liao2024maptrv2}. In multi-structure VE, vectors differ in geometry and point count, complicating shape and topology optimization. We introduce key-point dynamic matching to impose shape-specific constraints. After instance-level pairing \cite{carion2020end, zhu2020deformable}, we solve a point-wise bipartite matching between predicted vectors $\hat{P}={\left\{ \hat{p_i} \right\}}^M_{i=1}$ and ground truth $P={\left\{p_i \right\}}^T_{i=1}$, where $M$ is the fixed number of predicted points and $T$ varies with shape. Let $\beta$ denote point pairings and $\hat{C}={\left\{ \hat{c_i} \right\}}^M_{n=1}$ the predicted key-point probabilities, which are incorporated into the matching loss:
\begin{equation}
	\begin{aligned}
         \mathcal{L}_{match}(\hat{P}, P, \beta) &= \frac{1}{T} \sum_{i=1}^T(\alpha_{p} \cdot l_1(p_i, \hat{p_i}) + \alpha_{c} \cdot l_1(c_i, \hat{c_i})),
	\end{aligned}
\end{equation}
where $l_1$ denotes the $l_1$ loss. $\alpha_{p}$ and $\alpha_{c}$ are the balancing factors in the matching cost. The proposed DSC searches for the optimal $\beta^*$ with the lowest sequence matching cost:
\begin{equation}
	\begin{aligned}
    \beta^* & = \textrm{arg} \underset{\beta}{\min} \ \mathcal{L}_{match}(\hat{P}, P, \beta).
	\end{aligned}
\end{equation}
From the matching results, we extract the key-point sequence $\hat{P^k}$ from $\hat{P}$ for loss computation, allowing each ground-truth point to supervise its nearest prediction and reducing supervision misalignment.

\textbf{Vector Shape Supervision.}
To comprehensively constrain predicted vector shapes, we supervise global structure, local points, and key-point classification. Based on the matching results, the predicted key-point sequence $\hat{P^k}$ corresponds one-to-one with the ground truth $P$ of length $T$. The overall structure is measured using the average direction loss, preserving relative key-point positions, defined as:
\begin{equation}
	\begin{aligned}
         \mathcal{L}_{dir} &= \frac{1}{T} \sum_{i=1}^T \textrm{Cos\_ similarity( $\hat{d}^k_i$, $d_i$)},
	\end{aligned}
\end{equation}
where $\hat{d}^k_i$ and $d_i$ denote the $i$-th edge in the prediction and ground truth, respectively. Cosine similarity is calculated for each pair of edges. Subsequently, we use the $l_1$ loss to constrain the positional difference between the paired points, and the local point loss is expressed as:
\begin{equation}
	\begin{aligned}
         \mathcal{L}_{kp} &= \frac{1}{T} \sum_{i=1}^T ||\hat{p}^k_i - p_i||_1.
	\end{aligned}
\end{equation}
To model the dynamic key point number, a binary cross-entropy loss is adopted to supervise the probability of a predicted point being a key point. The calculation is as follows:
\begin{equation}
	\begin{aligned}
         \mathcal{L}_{cls} &= \frac{1}{M} \sum_{i=1}^M (\hat{c}_i, \mathbbm{1}_{\hat{p}_i \in \hat{p}^k}),
	\end{aligned}
\end{equation}
where $N$ is the predefined maximum number of points in a vector. $\mathbbm{1}_{A}$ is an indicator function which returns 1 if $A$ is true, and returns 0 otherwise. In summary, the vector shape loss is formulated as follows:
\begin{equation}
	\begin{aligned}
         \mathcal{L}_{VSL} &= \alpha_{1} \cdot \mathcal{L}_{dir} + \alpha_{2} \cdot \mathcal{L}_{kp} + \alpha_{3} \cdot \mathcal{L}_{cls},
	\end{aligned}
\end{equation}
where $\alpha_{1}$, $\alpha_{2}$ and $\alpha_{3}$ denote the weighted factors.

\section{Experiments}
We evaluate UniVector on both specific-structure and multi-structure VE tasks. First, we compare its performance on the Multi-Vector dataset for multi-structure VE. Next, we assess its results on existing specific-structure datasets. Finally, we conduct ablation studies to validate the proposed framework.

\subsection{Dataset and Implementation Details}

We present Multi-Vector, the first multi-structure vector extraction dataset, covering diverse categories and vector types. It contains 20,000 training and 3,734 validation images across three semantic categories: buildings, road boundaries, and center lines. Unlike existing datasets that focus on a single vector type, Multi-Vector includes polygons, polylines, and line segments. Leveraging building data from CrowdAI \cite{crowdAIMappingChallengeBaseline2018}, we re-annotated road boundaries and center lines commonly used in vector maps. All vectors are represented as directed point sequences in COCO \cite{lin2014microsoft} format. The dataset distribution is 70.6\% buildings, 18.9\% road boundaries, and 10.5\% center lines, with buildings as polygons, road boundaries as polylines, and center lines as line segments. This design better reflects practical applications and poses greater challenges than prior datasets. For more dataset details, please refer to the supplementary material.

To evaluate performance across vector types, we conduct structure-specific assessments. For buildings, we use mAP, IoU, CIoU, and PoLiS as \cite{zorzi2022polyworldg}. For road boundaries and center lines, we employ two levels of metrics: pixel-level (precision, recall, F1 with 10-pixel tolerance) and geometry-level, including Entropy-based Connectivity Metric (ECM) and Average Path Length Similarity (APLS)  as \cite{hu2024polyroad}.

\subsubsection{Specific-structure Datasets}
\textbf{CrowdAI} \cite{crowdAIMappingChallengeBaseline2018} contains over 280k training and 60k test images for building instance segmentation. The evaluation criteria are consistent with previous work \cite{zorzi2022polyworldg}, including COCO metrics, boundary mAP, CIoU, and PoLiS.

\textbf{Structured3D} \cite{zheng2020structured3d} is a synthetic 3D house dataset with projected top-view images, evaluated with room-, corner-, and angle-level precision, recall, and F1 scores \cite{yue2023connecting}.

\textbf{Topo-Boundary} \cite{xu2021topoboundary} includes 25k aerial images for road boundary extraction, assessed using pixel-level metrics with multiple tolerances and geometry-level metrics (ECM and APLS) \cite{hu2024polyroad}.

\textbf{Wireframe} \cite{huang2018learning} and \textbf{York Urban} \cite{denis2008efficient} are standard line segment detection datasets, evaluated using sAP and sF metrics at 10- and 15-pixel thresholds \cite{xu2021line}.

\subsubsection{Experiment Settings}
For the Multi-Vector dataset, we set 50 vector instances per image and 40 points per vector, using ResNet50 \cite{he2016deep} as the backbone and AdamW optimizer with a batch size of 6. Models are trained on 4 RTX 3090 GPUs for 30 epochs with an initial learning rate of $1 \times 10^{-4}$, decayed at epoch 27. Dynamic shape constraint parameters and loss weights follow empirically tuned values (see \cite{liao2024maptrv2}), and detailed settings for other datasets are provided in the code repository.

\subsection{Comparison with State-of-the-Art Methods}
\subsubsection{Multi-structure Vector Extraction}
We evaluate UniVector on the Multi-Vector dataset against representative specific-structure VE methods (Tables \ref{table_1}), showing that instance-geometry interaction improves geometric accuracy for buildings and other vector types, while simultaneously extracting multiple vector structures more efficiently. UniVector achieves top performance across most metrics, with 2–20× faster training and inference than cascaded multi-model approaches, and qualitative results confirm more accurate shapes and fewer false detections compared to prior methods; related experimental data are provided in the supplementary material.

\subsubsection{Specific-structure Vector Extraction}
\textbf{Polygon Extraction.}
Due to space limitations, only the CrowdAI results are presented in Table \ref{table_2}, where UniVector achieves state-of-the-art polygonal vector extraction, outperforming PolyR-CNN in AP/AR and surpassing RoomFormer in room-level metrics while remaining end-to-end. Visual comparisons show cleaner shapes and fewer false positives than previous methods, confirming UniVector’s higher geometric fidelity and robustness. Further experimental details and details are provided in the supplementary material.

\begin{table*}[]
\centering
\caption{Experimental results of different vector extraction methods on the Multi-Vector validation set. The results of Sat2Graph \cite{he2020sat2graph}, RNGDet++ \cite{xu2023rngdet++}, and SAM-Road \cite{hetang2024segment} on road boundaries and cente rlines are obtained through separate training.}
\resizebox{1.0\textwidth}{!}{
\begin{tabular}{cccccccccccccccc}
\toprule[1pt]
                                  &                                     & \multicolumn{4}{c}{\textbf{Building (polygon)}}               & \multicolumn{5}{c}{\textbf{Road   Boundary (polyline, line Segment)}}                                    & \multicolumn{5}{c}{\textbf{Center Line   (polyline, line Segment)}}                                       \\ \cmidrule(lr){3-6} \cmidrule(lr){7-11} \cmidrule(lr){12-16}  
\multirow{-2}{*}{\textbf{Method}} & \multirow{-2}{*}{\textbf{Backbone}} & \textbf{mAP}$\uparrow$  & \textbf{IoU}$\uparrow$  & \textbf{CIoU}$\uparrow$ & \textbf{PoLiS}$\downarrow$ & \textbf{Pre.}$\uparrow$ & \textbf{Rec.}$\uparrow$ & \textbf{F1-score}$\uparrow$ & \textbf{ECM}$\uparrow$  & \textbf{APLS}$\uparrow$ & \textbf{Pre.}$\uparrow$ & \textbf{Rec.}$\uparrow$ & \textbf{F1-score}$\uparrow$ & \textbf{ECM}$\uparrow$  & \textbf{APLS}$\uparrow$ \\ \midrule[0.5pt]
FFL \cite{girard2021polygonal}                               & ResNet-50                           & 44.5          & 76.2          & 56.4          & 2.89           & —             & —             & —                & —             & —                                     & —             & —             & —                & —             & —                                     \\
HiSup \cite{xu2022accurate}                             & ResNet-50                           & 45.3          & 77.5          & 58.2          & 2.56           & —             & —             & —                & —             & —                                     & —             & —             & —                & —             & —                                     \\
PolyR-CNN \cite{jiao2024polyrcnn}                         & ResNet-50                           & 48.3          & 77.2          & 56.4          & 2.41           & —             & —             & —                & —             & —                                     & —             & —             & —                & —             & —                                     \\
PolyR-CNN \cite{jiao2024polyrcnn}                        & Swin-L                              & 51.2          & 80.2          & 65.4          & 2.02           & —             & —             & —                & —             & —                                     & —             & —             & —                & —             & —                                     \\ \midrule[0.5pt]
Sat2Graph \cite{he2020sat2graph}                        & ResNet-50                           & —             & —             & —             & —              & 85.6          & 78.2          & 80.1             & 78.2          & 33.5                                  & 83.1          & 78.5          & 79.6             & 74.2          & 9.52                                  \\
RNGDet++ \cite{xu2023rngdet++}                         & ResNet-50                           & —             & —             & —             & —              & 84.7          & 92.9          & 87.1             & 83.3          & 40.3                                  & 82.2          & 92.3          & 86.1             & 79.2          & 12.2                                  \\
SAM-Road \cite{hetang2024segment}                          & VIT-B                               & —             & —             & —             & —              & 87.2          & 92.5          & 88.2             & 84.7          & 41.1                                  & 84.7          & 92.5          & 86.5             & 80.6          & 14.5                                  \\ \midrule[0.5pt] \rowcolor{gray!30}
UniVector                         & ResNet-50                           & 49.8          & 78.1          & 57.4          & 2.32           & 86.2          & \textbf{93.1}          & 88.4             & 85            & 42.1                                  & 84.3          & \textbf{95.5}          & 87.8             & 81.1          & 12.5                                  \\ \rowcolor{gray!30}
UniVector                         & Swin-L                              & \textbf{53.4} & \textbf{81.8} & \textbf{69.7} & \textbf{1.81}  & \textbf{90.0}   & 92.9 & \textbf{90.4}    & \textbf{88.9} & \textbf{47.8}                         & \textbf{88.4} & 90.4 & \textbf{88.2}    & \textbf{82.7} & \textbf{15.7}                         \\ \bottomrule[1pt]
\end{tabular}}
\label{table_1}
\end{table*}

\begin{table*}[]
 \centering
\caption{Experimental results of polygon extraction methods on the CrowdAI validation set. *Indicates the results from our retraining.}
\resizebox{1.0\textwidth}{!}{
\begin{tabular}{cccccccccccc}
\toprule[1pt]
\textbf{Method} & \textbf{Backbone} & \textbf{AP}$\uparrow$   & $\bm{\mathrm{AP}_{50}}$$\uparrow$ & $\bm{\mathrm{AP}_{75}}$$\uparrow$ & \textbf{AR}$\uparrow$   & $\bm{\mathrm{AR}_{50}}$$\uparrow$ & $\bm{\mathrm{AR}_{75}}$$\uparrow$ & $\bm{\mathrm{AP}_{boundary}}$$\uparrow$ & \textbf{IoU}$\uparrow$  & \textbf{CIoU}$\uparrow$ & \textbf{PoLiS}$\downarrow$ \\ \midrule[0.5pt]
Mask R-CNN \cite{he2017mask}      & ResNet-50         & 41.9          & 67.5          & 48.8          & 47.6          & 70.8          & 55.5          & 15.4                  & 61.3          & 50.1          & 3.45          \\
FFL \cite{girard2021polygonal}            & UResNet101        & 67.0            & 92.1          & 75.6          & 73.2          & 93.5          & 81.1          & 34.4                  & 84.3          & 73.8          & 1.95          \\
HiSup* \cite{xu2022accurate}          & HRNetV2-W48       & 64.7          & 86.5          & 74.6          & 67.6          & 87.6          & 76.8          & 39.9                 & 87.5         & 80.8         & 1.55           \\
PolyWorld \cite{zorzi2022polyworldg}      & R2U-Net           & 63.3          & 88.6          & 70.5          & 75.4          & 93.5          & 83.1          & 50.0                    & 91.2          & 88.3          & \textbf{0.96} \\
Re:PolyWorld \cite{zorzi2023re}   & —                 & 67.2          & 89.8          & 75.8          & —             & —             & —             & —                     & 92.2          & \textbf{89.7} & —              \\
GraphMapper \cite{wang2023regularized}    & —                 & 72.8          & 89.1          & 79.7          & 83.1          & 93.3          & 88.1          & —                     & 93.9          & 88.8          & —              \\
P2PFormer \cite{zhang2024p2pformer}       & ResNet-50         & 66.0            & 91.1          & 77.0            & —             & —             & —             & —                     & —             & —             & —              \\
P2PFormer \cite{zhang2024p2pformer}      & Swin-L            & 78.3          & 94.6          & 87.3          & —             & —             & —             & —                     & —             & —             & —              \\
PolyR-CNN \cite{jiao2024polyrcnn}      & ResNet-50         & 71.1          & 93.8          & 82.9          & 78.6          & 95.6          & 88.3          & 50.0                    & —             & —             & 1.57           \\
PolyR-CNN \cite{jiao2024polyrcnn}      & Swin-B            & 79.2          & 97.4          & 90.0            & 85.2          & 98.1          & 93.5          & 63.3                  & 91.6          & —             & 1.20            \\ \midrule[0.5pt] \rowcolor{gray!30}
UniVector       & ResNet-50         & 72.8          & 94.4          & 84.8          & 79.1          & 96.1          & 89.5          & 51.2                  & 92            & 88.2          & 1.34           \\ \rowcolor{gray!30}
UniVector       & Swin-B            & \textbf{79.9} & \textbf{98.2} & \textbf{90.8} & \textbf{86.3} & \textbf{98.9} & \textbf{94.2} & \textbf{64.2}         & \textbf{94.2} & 88.8          & 1.13           \\ \bottomrule[1pt]
\end{tabular}
}
\label{table_2}
\end{table*}

\textbf{Polyline Extraction.}
UniVector achieves near-SOTA polyline extraction on the Topo-Boundary dataset (Table \ref{table_3}), showing clear geometric advantages over both segmentation- and point-prediction methods, including large ECM/APLS gains versus OrientationRefine and higher accuracy than Enhanced-iCurb and RNGDet++ while maintaining faster inference (see supporting materials). Qualitative results further highlight UniVector’s smooth, topologically consistent road boundaries compared with the disconnections or coarse corners seen in competing methods. Further experimental details and details are provided in the supplementary material.

\begin{table}[]
\centering
\caption{Experimental results of polyline extraction methods on the Topo-Boundary validation set.}
\resizebox{0.8\textwidth}{!}{
\begin{tabular}{cccccc}
\toprule[1pt]
\textbf{Method}    & \textbf{Precision}$\uparrow$ & \textbf{Recall}$\uparrow$ & \textbf{F1-score}$\uparrow$ & \textbf{ECM}$\uparrow$  & \textbf{APLS}$\uparrow$ \\ \midrule[0.5pt]
OrientationRefine \cite{batra2019improved}  & 91.3               & 88.4            & 88.8              & 75.6          & 75.0            \\
RoadTracer \cite{bastani2018roadtracer}         & 79.1               & 82.1            & 79.8              & 82.4          & 73.9          \\
ConvBoundary \cite{liang2019convolutional}       & \textbf{93.4}      & 75.2            & 80.5              & 78.6          & 67.1          \\
VecRoad \cite{tan2020vecroad}           & 85.1               & 83.0              & 83.7              & 84.6          & 75.6          \\
iCurb \cite{xu2021icurb}              & 89.0                 & 87.3            & 87.7              & 88.9          & 82.6          \\
Enhanced-iCurb \cite{xu2021topoboundary}     & 89.4               & 86.4            & 87.4              & 89.3          & 82.2          \\
RNGDet \cite{xu2022rngdet}             & 87.9               & 87.6            & 88.3              & 88.5          & 82.1          \\
RNGDet++ \cite{xu2023rngdet++}          & 88.9               & 88.7            & 88.7              & 89.0            & 82.3          \\ 
PolyRoad \cite{hu2024polyroad}         & 91.6               & 88.6            & 89.2              & 89.5          & 82.8          \\ \midrule[0.5pt] \rowcolor{gray!30}
Univector & 91.6               & \textbf{89.1}   & \textbf{90.3}     & \textbf{89.9} & \textbf{83.2} \\ \bottomrule[1pt]
\end{tabular}}
\label{table_3}
\end{table}

\begin{table*}[]
\centering
\caption{Experimental results of line segment extraction methods on the Wireframe and York Urban validation sets.}
\resizebox{1.0\textwidth}{!}{
\begin{tabular}{cccccccccc}
\toprule[1pt]
\multirow{2}{*}{\textbf{Method}} & \multirow{2}{*}{\textbf{Epochs}} & \multicolumn{4}{c}{\textbf{Wireframe}}                                                                                                          & \multicolumn{4}{c}{\textbf{York Urban}}                                                                                                         \\ \cmidrule(lr){3-6} \cmidrule(lr){7-10}
                                 &                                  & \multicolumn{1}{l}{$\bm{\mathrm{sAP}^{10}}$$\uparrow$} & \multicolumn{1}{l}{$\bm{\mathrm{sAP}^{15}}$$\uparrow$} & \multicolumn{1}{l}{$\bm{\mathrm{sF}^{10}}$$\uparrow$} & \multicolumn{1}{l}{$\bm{\mathrm{sF}^{15}}$$\uparrow$} & \multicolumn{1}{l}{$\bm{\mathrm{sAP}^{10}}$$\uparrow$} & \multicolumn{1}{l}{$\bm{\mathrm{sAP}^{15}}$$\uparrow$} & \multicolumn{1}{l}{$\bm{\mathrm{sF}^{10}}$$\uparrow$} & \multicolumn{1}{l}{$\bm{\mathrm{sF}^{15}}$$\uparrow$} \\ \midrule[0.5pt]
DWP \cite{huang2018learning}                              & 120                              & 5.1                                & 5.9                                & —                                 & —                                 & 2.1                                & 2.6                                & —                                 & —                                 \\
HAWP \cite{xue2020holistically}                            & 30                               & 66.5                               & 68.2                               & 64.9                              & 65.9                              & 28.5                               & 29.7                               & 39.7                              & 40.5                              \\
LETR \cite{xu2021line}                        & 825                              & 65.2                               & 67.7                               & 65.8                              & 67.1                              & 29.4                               & 31.7                               & 40.1                              & 41.8                              \\
ULSD \cite{li2021ulsd}                             & 30                               & 68.8                               & 70.4                               & —                                 & —                                 & 28.8                               & 30.6                               & —                                 & —                                 \\
Re:PolyWorld \cite{zorzi2023re}                     & —                                & 50.2                               & 64.6                               & —                                 & —                                 & —                                  & —                                  & —                                 & —                                 \\
HAWPv2 \cite{xue2023holistically}                         & 30                               & 69.7                               & 71.3                               & —                                 & —                                 & 31.2                               & 32.6                               & —                                 & —                                 \\
PLNet \cite{xu2025airslam}                            & 40                               & 69.2                               & 70.9                               & —                                 & —                                 & 32                                 & 33.5                               & —                                 & —                                 \\ \midrule[0.5pt] \rowcolor{gray!30}
UniVector-R50                    & 30                               & 64.5                               & 66.5                               & 69.1                              & 69.9                              & 28.6                               & 30.8                               & 39.7                              & 40.5                              \\ \rowcolor{gray!30}
UniVector-SwinL                  & 30                               & \textbf{69.8}                      & \textbf{71.7}                      & \textbf{71.4}                     & \textbf{72.2}                     & \textbf{33.2}                      & \textbf{35.1}                      & \textbf{44.5}                     & \textbf{45.8} \\ \bottomrule[1pt]                  
\end{tabular}
}
\label{table_4}
\end{table*}

\textbf{Line Segment Extraction.}
UniVector delivers the highest accuracy on Wireframe and York Urban (Table \ref{table_4}), slightly exceeding PLNet in $\mathrm{sAP}{10}$ and showing larger gains in $\mathrm{sF}{10}$, with strong cross-domain generalization (see supporting materials). Qualitative results further demonstrate cleaner, more reliable line detection than L-CNN, HAWP, or LETR, reducing false or noisy segments. Further experimental details and details are provided in the supplementary material.

\begin{figure*}[t]
	\centering
	\includegraphics[width=0.95\textwidth]{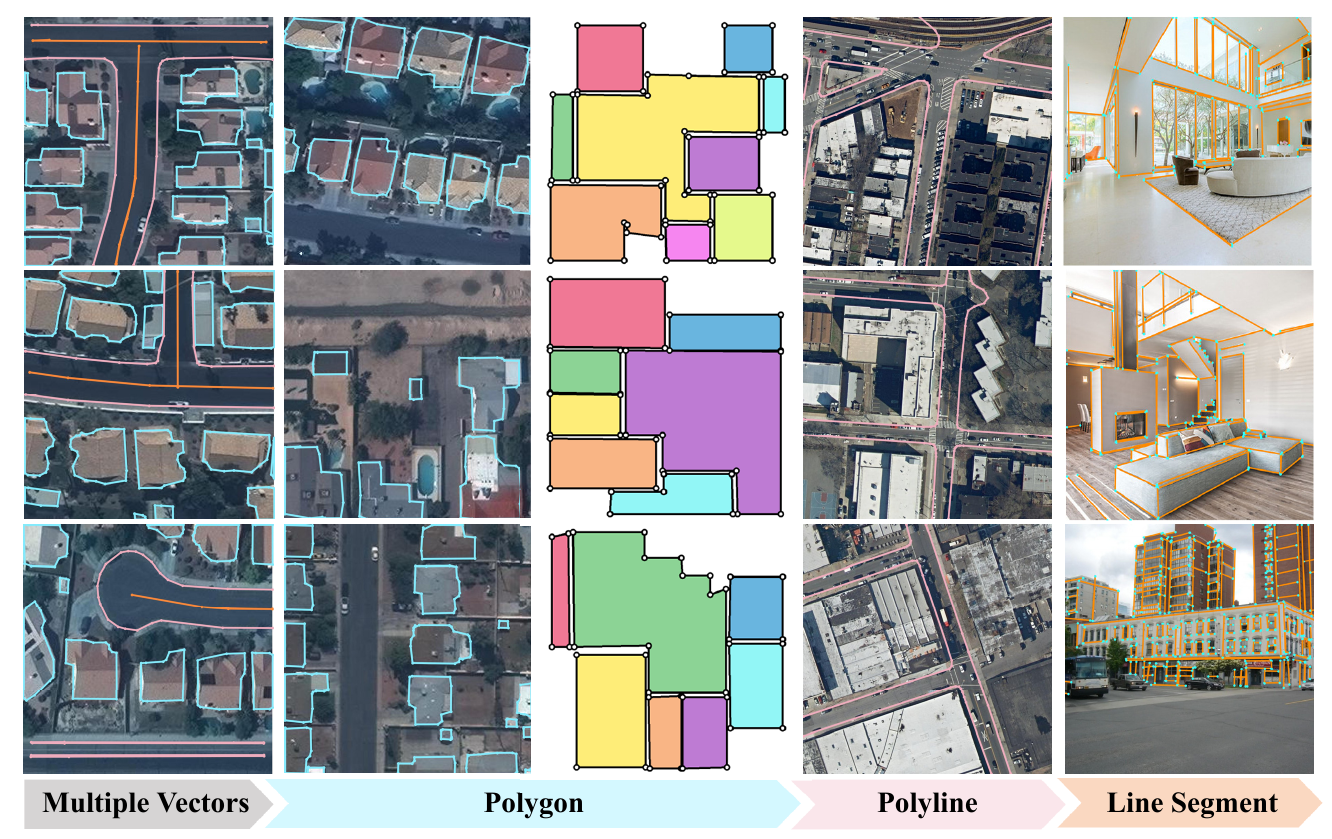}
	\caption{Visualization of UniVector on different datasets, including polygons, polylines, line segments, and the simultaneous extraction of all three.}
	\label{fig_5}
\end{figure*}
\subsection{Ablation Studies}
We perform ablation studies using a ResNet-50 backbone and 30-epoch training to assess each UniVector component, analyzing module design and hyperparameter choices.
Performance is evaluated with mAP for buildings and F1-score for road boundaries and center lines; further details appear in the supporting materials.

\subsubsection{Ablation Study on UniVector}
We perform ablation on multi- and single-structure datasets using RoomFormer’s geometry-only decoding as the baseline. As shown in Table \ref{table_5}, Instance-Geometry Interaction Decoding (IGID) provides the largest gains, while Unified Vector Encoding (UVE) and Dynamic Shape Constraint (DSC) offer additional improvements in query initialization and training supervision, especially for complex buildings and road boundaries.

\subsubsection{Discussion of Unified Vector Encoding}

\textbf{Comparison of Different Encoding Methods.} We compare random encoding \cite{carion2020end}, hierarchical encoding \cite{liao2024maptrv2}, and our UVE. Random encoding yields disordered vectors, hierarchical encoding captures only positional cues, whereas UVE integrates instance detection with geometric deformation for richer, geometry-aware initialization. Please refer to the supplementary materials for related visualization results.

\begin{table*}[]
\centering
\caption{Ablation studies on the three modules of UniVector across datasets Multi-Vector, CroadAI, Topo-Boundary, and Wireframe.}
\resizebox{1\textwidth}{!}{
\begin{tabular}{cccccccccc}
\toprule[1pt]
\multirow{2}{*}{\textbf{Baseline}} & \multirow{2}{*}{\textbf{IGID}} & \multirow{2}{*}{\textbf{UVE}} & \multirow{2}{*}{\textbf{DSC}} & \multicolumn{3}{c}{\textbf{Multi-Vector}}                                  & \multirow{2}{*}{\textbf{CrowdAI}} & \multirow{2}{*}{\textbf{Topo-Boundary}} & \multirow{2}{*}{\textbf{Wireframe}} \\  \cmidrule{5-7}
                          &                       &                      &                      & \textbf{Building}             & \textbf{Road Boundary}       & \textbf{Center Line}          &                          &                                &                            \\ \midrule[0.5pt]
\checkmark                         & \multicolumn{1}{l}{}  & \multicolumn{1}{l}{} & \multicolumn{1}{l}{} & 39.6                 & 77.2                & 78.3                 & 63.9                     & 78.8                           & 62.3                       \\
\checkmark                         & \checkmark                     &                      &                      & 45.2 (+5.6)          & 83.2 (+6.0)         & 83.8 (+5.5)          & 69.3 (+5.4)              & 85.6 (+6.8)                    & 66.8 (+4.5)                \\
\checkmark                         & \checkmark                     & \checkmark                    &                      & 47.6 (+2.4)          & 85.4 (+2.2)         & 86.3 (+2.5)          & 71.5 (+2.2)              & 87.5 (+1.9)                    & 68.2 (+1.4)                \\ \rowcolor{gray!30}
\checkmark                         & \checkmark                     & \checkmark                    & \checkmark                    & \textbf{49.4 (+1.8)} & \textbf{87.8 (+2.4)} & \textbf{88.6 (+2.3)} & \textbf{72.8 (+1.3)}     & \textbf{90.3 (+2.8)}           & \textbf{69.1 (+0.9)}  \\ \bottomrule[1pt]     
\end{tabular}
    }
\label{table_5}
\end{table*}

\subsubsection{Discussion of Instance-Geometry Interaction}
\textbf{What Have Structured Queries Learned?}
We visualized decoder attention maps across different layers to verify the effectiveness of the structured queries, with the experimental data and visualization results provided in the supplementary materials. Decoder attention maps reveal that instance queries capture global structures while geometry queries focus on local details, and their iterative cross-layer interactions progressively refine reference points, demonstrating that instance-geometry interaction significantly enhances vector extraction accuracy.

\textbf{How to Implement Instance-Geometry Interaction?}
To validate the effectiveness of instance–geometry interaction, we experimented with different interaction strategies. Ablation experiments show that instance-geometry (I-G) and instance-level (I-I) interactions significantly boost accuracy with minimal overhead, whereas geometry-only (G-G) and global (Full) interactions yield weaker performance, with Full incurring about 40 \% extra cost from cross-instance interference (data and visualizations are provided in the supplementary material).

\subsubsection{Discussion of Dynamic Shape Constraint}
Ablation studies on the dynamic shape constraint (DSC) show that removing it or using only smooth $l_1$ or directional loss reduces performance, while combining keypoint loss $L_{kp}$ and directional loss $L_{dir}$ with an optimal weight ratio of 10:1 yields the best results.

Ablation studies reveal that removing the dynamic shape constraint or using only smooth $l_1$ or directional loss lowers performance, whereas combining keypoint loss $L_{kp}$ and directional loss $L_{dir}$ at a 10:1 ratio delivers the best results (data and visualizations are provided in the supplementary material).

\section{Conclusion}
We propose UniVector, a unified framework for simultaneously extracting multiple vector structures—including polygons, polylines, and line segments—by encoding them into a shared representation and refining their positions and shapes through instance-geometry interaction. To evaluate its performance on complex multi-structure scenes, we construct the Multi-Vector dataset from CrowdAI, covering polygons, polylines, and line segments. Experiments show that UniVector achieves state-of-the-art results on both traditional single-structure and more challenging multi-structure VE tasks. Future work will focus on developing a zero-shot vector extraction foundation model and applying vector representations to downstream tasks such as visual localization and path planning.

\section*{Acknowledgment}
This work was supported in part by the National Natural Science Foundation of China under National Science Fund for Distinguished Young Scholars 62425109, and Grant U22B2014; and in part by the Science and Technology Plan Project Fund of Hunan Province under Grant 2022RC3064.

\bibliographystyle{elsarticle-num-names}

\bibliography{ref}

\end{document}